\documentclass[11pt]{article}

\usepackage[final]{latex-simple}

\usepackage{natbib}

\usepackage{amsmath,amsfonts,bm}

\def\eqref#1{equation~\ref{#1}}

\def\1{\bm{1}}

\DeclareMathAlphabet{\mathsfit}{\encodingdefault}{\sfdefault}{m}{sl}
\SetMathAlphabet{\mathsfit}{bold}{\encodingdefault}{\sfdefault}{bx}{n}

\usepackage[utf8]{inputenc} 
\usepackage[T1]{fontenc}    
\usepackage{hyperref}       
\usepackage[dvipsnames]{xcolor}
\hypersetup{
    colorlinks=true,
    urlcolor=Orchid,
    linkcolor=DarkOrchid,
    citecolor=DarkOrchid,
    }
\usepackage{url}            
\usepackage{booktabs}       
\usepackage{amsmath}
\usepackage{amssymb}
\usepackage{mathtools}
\usepackage{amsthm}
\usepackage{amsfonts}       
\usepackage{nicefrac}       
\usepackage{microtype}      
\usepackage{xcolor}         
\usepackage{blindtext}
\usepackage[capitalise]{cleveref}
\usepackage{graphicx}
\usepackage{subcaption}
\usepackage{microtype} 
\usepackage{comment}
\usepackage{siunitx}
\DeclareSIUnit\angstrom{\text{Å}}
\usepackage{tikz}
\usetikzlibrary{positioning,calc}
\usepackage{adjustbox}

\newcommand{\cbrace}[1]{\left\{#1 \right\}} 
\newcommand{\pr}[1]{\left(#1 \right)} 

\theoremstyle{plain}

\theoremstyle{definition}

\theoremstyle{remark}

\title{Crystal-GFN: sampling crystals with desirable properties and constraints}

\author[0]{Mila AI4Science}
\author[1,2]{\nameemail{Alex Hernandez-Garcia}{alex.hernandez-garcia@mila.quebec}}
\author[1,3]{\nameemail{Alexandre Duval}{alexandre.duval@mila.quebec}}
\author[1,2]{\nameemail{Alexandra Volokhova}{alexandra.volokhova@mila.quebec}}
\author[1,2]{\nameemail{Yoshua Bengio}{yoshua.bengio@mila.quebec}}
\author[1,4]{\nameemail{Divya Sharma}{dsharm23@jh.edu}}
\author[1]{\nameemail{Pierre Luc Carrier}{pierre.luc.carrier@mila.quebec}}
\author[2,5]{\nameemail{Yasmine Benabed}{benabed.yasmine@hydroquebec.com}}
\author[1,2]{\nameemail{Michał Koziarski}{michal.koziarski@mila.quebec}}
\author[1,2]{\nameemail{Victor Schmidt}{schmidtv@mila.quebec}}
\author[6]{\nameemail{Gian-Marco Rignanese}{gian-marco.rignanese@uclouvain.be}}
\author[1,6]{\nameemail{Pierre-Paul De Breuck}{pierre-paul.de-breuck@mila.quebec}}
\author[4]{\nameemail{Paulette Clancy}{pclancy3@jhu.edu}}

\affil[1]{Mila, Quebec AI Institute}
\affil[2]{Université de Montréal}
\affil[3]{CentraleSupélec, Université Paris-Saclay}
\affil[4]{Johns Hopkins University}
\affil[5]{Centre d'excellence en électrification des transports et stockage d'énergie, Hydro-Québec}
\affil[6]{UCLouvain}
\affil[0]{This team name is the way the authors have found to express that the customary ordered list of authors hardly reflects their contributions. All authors have actively and significantly contributed to this work. The list in this manuscript has been randomised. \hfill \break This is the version of the manuscript submitted---though not accepted---to \href{https://openreview.net/forum?id=cPMpWTV9Rt}{ICML 2024}. \hfill \break Correspondence email: \nameemail{alex.hernandez-garcia@mila.quebec}{alex.hernandez-garcia@mila.quebec}}

\hypersetup{%
  pdfauthor={Mila AI4Science, Alex, Alexandre, Alexandra, Yoshua, Divya, Pierre Luc Michał, Victor}, 
  pdftitle={Crystal-GFN: sampling crystals with desirable properties and constraints},
  pdfsubject={Crystal-GFN},
  pdfkeywords={crystals,gflownet}
}

\begin{document}

\maketitle

\begin{abstract}
    The discovery of novel solid-state materials, such as electrocatalysts, super-ionic conductors, or photovoltaic materials, plays a critical role in addressing various global challenges. It has, for instance, the potential to significantly improve the efficiency of renewable energy production and storage, thereby making substantial contributions to climate crisis mitigation strategies.
    In this paper, we introduce Crystal-GFN, a generative model of crystal structures possessing desirable properties and constraints. Operating as a multi-environment, continuous-discrete GFlowNet, it sequentially samples structural attributes of crystalline materials, namely space group, composition and lattice parameters. This domain-inspired approach enables the flexible incorporation of physicochemical and geometric hard constraints. 
    We demonstrate the capabilities of Crystal-GFN to efficiently discover diverse and valid crystals with various properties: low predicted formation energy (median -3.2 eV/atom), band gap close to a target value and high density.
    Overall, Crystal-GFN is a crystal generation method that addresses several existing challenges in the literature and opens promising paths for accelerating materials discovery with machine learning.


\end{abstract}


\section{Introduction}
\label{sec:introduction}

\begin{figure*}[t]
    \centering
    \begin{adjustbox}{width=\textwidth}
      \input{tikz/crystalgfn.tex}
    \end{adjustbox}
    \caption{A schematic of the crystal generation process of Crystal-GFN. First, the space group is selected, which may be sampled directly or after the selection of the crystal-lattice system and/or the point symmetry. 
    Then, a composition is generated by iteratively selecting an element type and its quantity. Finally, the lattice parameters of the unit cell are sampled. We introduce hard constraints, denoted by $C_i$, within and between these components, as described in \cref{sec:methods}.}
    \label{fig:crystalgfn}
\end{figure*}

Materials discovery plays an essential role in tackling the global warming challenge \citep{ipcc2023}, particularly by transforming industries responsible for a significant fraction of the anthropocentric greenhouse gas emissions.
From developing cutting-edge photovoltaic cells or new-generation solid-state batteries to addressing carbon capture or catalysis, the quest for innovative materials with targeted properties has the power to reshape the technological landscape~\citep{zitnick2020introduction, pyzer2022accelerating}.

However, the discovery of new solid-state materials faces major challenges, encompassing crystal structure generation \cite{ren2022invertible}, property prediction \citep{duval2023faenet} and synthesis \citep{szymanski2023autonomous}.
In practice, given the vast space of possible materials, 
domain knowledge is key to restrict the search space to fewer, more promising structures. 
Furthermore, due to the cost of synthesis and quantum mechanics simulation,
traditional trial-and-error processes are extremely costly\citep{oganov2019structure}. 
This provides  machine learning (ML) with the opportunity to greatly accelerate the generation and evaluation of promising candidates, broadening materials exploration by orders of magnitude \citep{merchant2023scaling, zeni2023mattergen}.

While ML has already made considerable advances in property prediction \cite{debrMaterials,liao2023equiformerv2, batatia2023foundation} and small molecule generation~\citep{pakornchote2023diffusion,zheng2023towards},  its impact in the field of generating solid-state materials lags behind.
At the core of this disparity lies the challenge of modeling the crystalline structure, characterised by a unit cell with symmetries that repeats in all three dimensions\footnote{See \Cref{sec:app:crystals} for formal definitions.}.
This periodicity makes it difficult to use finite graphs as representations for ML models.
Furthermore, discovering stable candidates requires exploring the combinatorial space of chemical compositions and all associated possible periodic arrangements, while respecting the complex physics that make them viable and (meta)-stable structures \citep{zhu2023wycryst, zhao2023physics,therrien2021metastability}. 
Indeed, crystal generation involves a multitude of factors, including specific bonding preferences between elements, geometric and chemical constraints, and a positioning that is correlated with the local energy minimum defined by quantum mechanics \citep{xie2021crystal}. 

Overall, existing methods often struggle to capture the above complexities in the generation process, failing to generate crystals that conform to fundamental crystallography principles, such as symmetry, and thus fail to consistently generate valid crystals \citep{court20203, long2021constrained, xie2021crystal, zeni2023mattergen}. 
Besides, many methods also show difficulty to discover diverse and novel crystals with specific properties \citep{zhao2023physics, merchant2023scaling, yang2023scalable}.

In this work, we introduce Crystal-GFN (\cref{fig:crystalgfn}), a generative model based on GFlowNets \cite{bengio2021gfn}, designed to sample inorganic crystal structures with desirable properties and constraints, in a space inspired by theoretical crystallography. Our domain-inspired approach sequentially selects a space group, a composition and lattice parameters to parameterise a crystal structure (\cref{sec:methods}), while enforcing strict physicochemical and structural constraints in the sampling process. 
These constraints are inspired by crystallography: the lattice parameters obey the relations defined by the crystal-lattice system; and the compositions satisfy multiplicities laid out by the space group.  
After training, Crystal-GFN samples crystal structures proportionally to the property used as a reward during training, hence providing diverse sampling. This key property enables an efficient exploration of the material space with respect to one or more properties of interest, estimated by any meaningful proxy model. In this case we trained two ML models of the formation energy (related to material stability) and of the band gap (electronical properties), as well as a fixed measure of unit cell's density. Note that our method can be flexibly extended, allowing for the inclusion of additional constraints and confining the sampling space to specific compositions or space groups of interest



In this context, it should be noted that we define a crystal as the union of composition, space group, and lattice parameters, excluding the species coordinates. This is sufficient to distinguish between most polymorphs and closely aligns with the data accessible to experimentalists. Nonetheless, if necessary, atomic positions can be reconstructed using conventional global optimization techniques~\cite{oganov2019structure} or ML force fields~\cite{chenUniversalGraphDeep2022}.

The Crystal-GFN models are trained in under 30 hours on a CPU-only machine. For evaluation, we sample 10,000 compounds from the trained Crystal-GFN to demonstrate its ability to sample diverse candidates with low formation energy, targeted band gap or high density. For example, 95 \% of the sampled structures have a predicted energy lower than -2.0 eV/atom and the overall median is -3.2 eV/atom. For reference, the median of the training data is less than -2 eV/atom.
Finally, we evaluate the conditional generation capabilities of our Crystal-GFN by applying restrictions at sampling time. For example, we limit the compositions to binary Fe-O or ternary Li-Mn-O, to a maximum number of atoms or to cubic lattices only.




\section{Related Work}
\label{sec:related-work}

There exists a record of generative methods for molecular and crystal structure generation, generally falling into one of these categories:
Variational Auto Encoders (VAE) \citep{ren2022invertible, pakornchote2023diffusion},
Generative Adversarial Networks (GANs) \citep{cao2018molgan, nouira2018crystalgan, ren2020inverse, long2021constrained},
Normalising flows \citep{satorras2021en, ahmad2022free}
and Diffusion models \citep{xu2022geodiff, zheng2023towards}. 
In general, these methods are trained to reconstruct the training data from a latent distribution by maximising its likelihood or minimising the discrepancy between generated samples and the real distribution. Once trained, they can generate new instances by sampling from the learned latent space. 
Importantly, the likelihood of obtaining a particular sample should be invariant to 3D euclidean transformations~\citep{duval2023hitchhiker}, ensuring symmetry invariance\footnote{Intuitively, the model should not have a preferred direction: a sample or a rotation of that sample should be equally likely.}. While this is not satisfied by all methods, like GANCSP \citep{kim2020generative} or CubicGAN \citep{zhao2021high}, Crystal-GFN produces invariant crystal representations by design.

Very recently, GNoMe \citep{merchant2023scaling} and MatterGen \citep{zeni2023mattergen} truly showed the potential of generative methods for inorganic crystals. Despite being comprehensive studies that included in-depth DFT calculations at a massive scale, these two methods still reflect common limitations of the field. 
MatterGen adopts a diffusion process on atom types, lattice parameters and atom coordinates, neglecting crystal symmetries despite being a major characteristic of periodic crystal structures. As pointed out by the authors, the generated crystals belong to P1, the least symmetric space group. Thus, MatterGen is very limited in the diversity of the explored space groups. This limitation extends to other diffusion models, like CDVAE \citep{xie2021crystal}, because diffusion does not (yet?) keep the symmetries of the structure intact. 
On the other hand, GNoME performs partial substitution or random search followed by relaxation using an ML potential to find new structures. This can limit the exploration space both in terms of diversity of composition and space groups. Besides, as most generative models, it does not allow to effectively "search" the material space for candidates with a specific functional property.


Overall, existing generative models for crystal structures struggle to embed essential physicochemical constraints, 
often leading to physically unrealistic and unstable structures. To tackle this, we incorporate domain knowledge from materials science and crystallography to sequentially build materials by selecting the space group, composition and lattice parameters while imposing hard compatiblity and validity constraints. Therefore, the crystal structures sampled by Crystal-GFN are always valid, resulting in more physically viable structures. One core advantage of using GFlowNets to generate crystals is their ability to discover diverse structures with a given target property (mode-mixing). Finally, there is no conceptual restriction in the number of distinct elements / space groups that we can incorporate.


\section{Background}
\label{sec:background}

In this section, we briefly review the necessary background on GFlowNets and crystallography, before describing our proposed method, Crystal-GFN. \Cref{sec:app:crystals} provides additional background about materials and crystal structures.

\subsection{GFlowNets}
\label{sec:gflownets}

Generative Flow Networks (GFlowNets or GFN for short), were introduced by \citet{bengio2021gfn} as an amortised inference method to sample from high-dimensional distributions, where traditional methods, such as MCMC and reinforcement learning are inefficient in terms of mode mixing. 
In essence, GFlowNets are learn a sampling policy $\pi$ to generate objects $x \in \mathcal{X}$ proportionally to a non-negative reward function $R(x)$, that is $\pi(x) \propto R(x)$. 
This facilitates the discovery of multiple modes of a given reward function, which is a desirable objective in scientific discovery ~\citep{jain2023scientificdiscovery}. 
For instance, GFlowNets have been successfully applied for biological sequence design and molecular property prediction as part of active learning algorithms \citep{jain2022bioseq,hg2023multifidelity}.

A key property of GFlowNets is that objects are generated sequentially. 
Starting from a special state $s_0$, transitions $s_{t} {\rightarrow} s_{t+1} \in \mathbb{A}$ are applied between states $s \in \mathcal{S}$, forming trajectories $\tau=(s_0 \rightarrow s_1 \rightarrow \ldots \rightarrow x)$, where $\mathbb{A}$ is a predefined action space and $\mathcal{S}$ the state space. 
This sequential construction of objects, together with the fact that the (forward) transition policy $P(s_{t+1}|s_t)$ is parameterised by a neural network (with parameters $\theta$), provides amortisation and the potential of systematic generalisation.
Generalisation is possible if the decomposition into partially constructed objects provides a structure that can be learned by a machine learning model. 

In order to obtain a policy that samples proportionally to the reward distribution, several training objectives have been proposed in the literature.
Here, we will make use of the Trajectory Balance objective \citep{malkin2022tb}, which has proven effective in various tasks \citep{jain2022bioseq,hg2023multifidelity,malkin2022tb}.

While GFlowNets were initially introduced for discrete probabilistic modelling, \citet{lahlou2023continuousgfn} recently proposed a generalisation of the framework for continuous or hybrid state spaces.
This work empirically demonstrates this generalisation, as we propose a GFlowNet that operates in a hybrid---mixture of discrete and continuous---state space.

\subsection{Crystals}
\label{sec:crystals}

Crystals are highly structured solid materials defined by a repeated arrangement of atoms in space. The elementary pattern that periodically repeats itself is called a unit cell.
This unique ordering constitutes the material's signature and greatly influences its properties. Hence, the understanding of crystal structures and the structure-property relationship is crucial in the development of new materials.
While there exists infinite crystal structures, the field of theoretical crystallography has developed ways to systematically parameterise and classify crystals according to their symmetry. 

\paragraph{Lattice} 
An $n$-dimensional \textit{lattice} $\Lambda$ can be defined as the set of integral combinations of the linearly independent \textit{non-coplanar lattice basis vectors} $\mathbf{a}_i \in \mathbb{R}^n$: $\Lambda \doteq \cbrace{\sum_i^n {m}_i \mathbf{a}_i \mid m_i \in \mathbb{Z}}$, given a reference origin. In 3D, the 3 lattice basis vectors can be converted into 6 parameters: $a, b, c$ determine the lengths of each basis vector and $\alpha, \beta, \gamma$ the angles between the basis vectors.

\paragraph{Composition}
We call composition the set of all atom types comprised in the crystal's unit cell, with their respective quantities.

\paragraph{Space group}
The specific arrangements of atoms in a crystal will generate some degree of symmetry. Symmetry is achieved when one or more operations exist that leave the initial arrangement of atoms unchanged. The combination of all present symmetry elements in a crystal indicates its space group.
A space group element can be described as a tuple $\pr{\mathbf{W}, \mathbf{t}}$, where $\mathbf{W}$ is the orthogonal matrix and $\mathbf{t}$ the translation vector. An element maps a vector $\mathbf{x} \in \mathbb{R}^n$ to $\mathbf{W} \mathbf{x} +\mathbf{t}$. 
In 3D, space groups are classified into 230 types. 


\section{Crystal-GFNs}
\label{sec:methods}

In this section, we describe the details of Crystal-GFN, the main method we introduce in this paper to explore the space of crystal structures and generate crystals with desirable properties and domain constraints. 

Drawing inspiration from crystallography, we represent crystals as the concatenation of three distinct components or \textit{subspaces}: space group (SG), composition (C) and lattice parameters (LP) of the unit cell.
The Cartesian product of the spaces of these three components would make the \textit{state space} $\mathcal{S} = \mathcal{S}_{SG} \times \mathcal{S}_{C} \times \mathcal{S}_{LP}$ and \textit{sample space} $\mathcal{X} = \mathcal{X}_{SG} \times \mathcal{X}_{C} \times \mathcal{X}_{LP}$ of a naive, unrestricted implementation with GFlowNets.
An important advantage of the GFlowNet framework and of the domain-inspired data representation that we propose is that it allows us to flexibly introduce domain knowledge such as well-studied geometrical constraints from crystallography as well as physicochemical constraints, described below in this section.
These constraints are mostly introduced by restricting the action space $\mathbb{A}$ of Crystal-GFN, both within and across subspaces.

In this work, we design the Crystal-GFN such that trajectories first select the space group, then the composition and finally the lattice parameters. 
\Cref{fig:crystalgfn} offers a graphical summary of the sequential generation in a Crystal-GFN.
Below we describe the details of the three subspaces and the constraints we have introduced.

\subsection{Space group}
\label{sec:spacegroup}

In our Crystal-GFN, the sample space for the space group subspace is $\mathcal{X}_{SG} = \{1, 2, \ldots,  230\}$, corresponding to the 230 symmetry groups in 3D.
As discussed in \cref{sec:gflownets}, GFlowNets rely on the decomposition of objects into multiple steps to facilitate generalisation. Therefore, we draw inspiration from theoretical crystallography to incorporate additional structure into the space group subspace.

A lattice carries 14 possible arrangements, referred to as the Bravais lattices, that are organised into seven \textit{crystal systems}: triclinic, monoclinic, orthorhombic, tetragonal, trigonal, hexagonal and cubic.
A related but slightly different categorisation is the \textit{lattice system}, with also seven systems (see \cref{fig:latticesystems}).
The lattice system category is convenient for our purposes because it imposes specific constraints on the lattice parameters (see $C_4$ in \cref{fig:crystalgfn}).
Here, we use a derived categorisation resulting from the combination of a crystal system and a lattice system.
We refer to it as \textit{crystal-lattice system} and it has 8 categories: triclinic, monoclinic, orthorhombic, tetragonal, trigonal-rhombohedral, trigonal-hexagonal, hexagonal-hexagonal and cubic.

Besides the crystal-lattice system, we introduce another category in the Crystal-GFN space group subspace: the \textit{point symmetry} (or site symmetry), which defines the type of symmetry of a point group.
We use the following 5 categories of point symmetries: centrosymmetric, non-centrosymmetric, enantiomorphic, polar and enantiomorphic-polar.
We choose these categories because the combination of point symmetry and crystal system gives rise to one of the 32 crystal classes or crystallographic point groups.

In sum, our space group subspace $\mathcal{S}_{SG}$ is a three-dimensional set where the entries correspond to the crystal-lattice system, the point symmetry and the space group. We define the action space such that, starting from the source state $s_0$, the transitions can set any of the three dimensions. However, the selection of an option along any dimension restricts the valid options for the remaining dimensions (constraint $C_1$ in \cref{fig:crystalgfn}). For example, if point symmetry ``non-centrosymmetric'' is selected from the source state, only three crystal-lattice systems remain valid and the number of valid space groups is reduced to 25. The ``stop'' action is valid if and only if the space group has been selected.

\subsection{Composition}
\label{sec:composition}

We represent a composition as the number of atoms of each element present in the unit cell.
Specifically, for a vocabulary of $D$ elements, where each element can have up to $K$ atoms, we construct $D$-dimensional vectors where each entry indicates the number of atoms of element $d$.
This yields the space $\mathcal{S}_C = \{(k_1, \ldots k_D) | k_d \in \{0, 1, \ldots K\}, d = 1, \ldots D\}$. 
The action space of the composition subspace consists of the choice of element and number of atoms, that is $\mathbb{A}_C = \{1, \ldots D\} \times \{1, \ldots K\}$, plus the special ``stop'' action to finish a trajectory.

To meet the electroneutrality requirement, the sum of positive and negative charges in a material (considering an ionic approximation of their bonds) must be equal.
Since we have control over the action space of the GFlowNet, we incorporate a hard constraint to ensure that all generated compositions have a neutral charge ($C_3$ in \cref{fig:crystalgfn}).
We argue this is an advantage over methods that aim at learning such properties implicitly from the data \citep{xie2021crystal}.

Finally, we also have an opportunity to incorporate inter-subspace constraints between the space group and the composition. Because of the symmetry imposed by each space group, not all compositions are possible given a space group, and given a composition not all space groups are valid. By way of illustration, the most restrictive space group---international number 230--cannot accommodate compositions with fewer than 16 atoms per element. Since here the trajectories of a Crystal-GFN first sample a space group, we restrict the action space of the GFlowNet to only allow the sampling of compositions that are compatible with the Wyckoff positions (see \cref{sec:app:crystals}) of the selected space group (constraint $C_2$ in \cref{fig:crystalgfn}). 

\subsection{Lattice parameters}
\label{sec:latticeparameters}

The lattice parameters ($a, b, c, \alpha, \beta, \gamma$), in combination with the lattice system (or crystal system) determine the shape of the unit cell in three dimensions. 
While both the composition and the space group subspaces are discrete, 
the lattice parameters are real-valued and thus yield a continuous sample space 
$\mathcal{X}_{LP} = \{(\ell_{min}, \ell_{max})^3 \times (\theta_{min}, \theta_{max})^3\}$
, where 
$\ell_{min}$ and $\ell_{max}$ 
are the minimum and maximum lengths and 
$\theta_{min}$ and $\theta_{max}$ 
are the minimum and maximum angles.
In order to sample sets of continuous lattice parameters, the GFlowNet policy model outputs the parameters of a six-dimensional mixture of Beta distributions, a continuous probability distribution defined in the interval $(0, 1)$. 

\begin{figure*}[htb]
  \centering
  \begin{subfigure}[b]{0.09\textwidth}
      \centering
			\includegraphics[width=\textwidth]{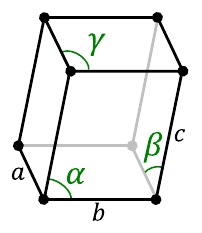}
			\caption{}
			\label{fig:triclinic}
  \end{subfigure}
  \hfill
  \begin{subfigure}[b]{0.09\textwidth}
      \centering
			\includegraphics[width=\textwidth]{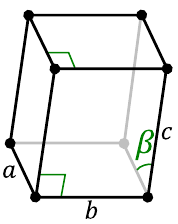}
			\caption{}
			\label{fig:monoclinic}
  \end{subfigure}
  \hfill
  \begin{subfigure}[b]{0.09\textwidth}
      \centering
			\includegraphics[width=\textwidth]{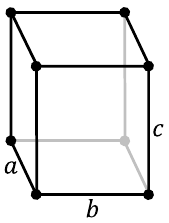}
			\caption{}
			\label{fig:orthorhombic}
  \end{subfigure}
  \hfill
  \begin{subfigure}[b]{0.09\textwidth}
      \centering
			\includegraphics[width=\textwidth]{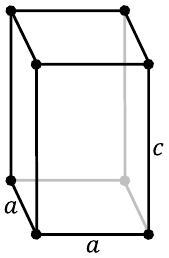}
			\caption{}
			\label{fig:tetragonal}
  \end{subfigure}
  \hfill
  \begin{subfigure}[b]{0.09\textwidth}
      \centering
			\includegraphics[width=\textwidth]{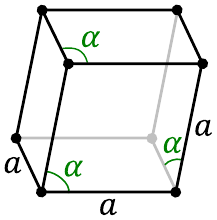}
			\caption{}
			\label{fig:rhombohedral}
  \end{subfigure}
  \hfill
  \begin{subfigure}[b]{0.09\textwidth}
      \centering
			\includegraphics[width=\textwidth]{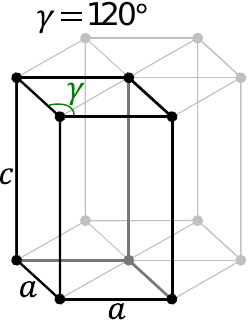}
			\caption{}
			\label{fig:hexagonal}
  \end{subfigure}
  \hfill
  \begin{subfigure}[b]{0.09\textwidth}
      \centering
			\includegraphics[width=\textwidth]{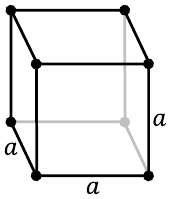}
			\caption{}
			\label{fig:cubic}
  \end{subfigure}
  \hfill
	\caption{The seven lattice systems and the constraints they impose on the lattice parameters. In order a--g: triclinic, monoclinic, orthorhombic, tetragonal, rhombohedral, hexagonal and cubic. From~\cite{wikicommons-lattice}}
	\label{fig:latticesystems}
\end{figure*}

Here, we incorporate an inter-subspace constraint from the space group ($C_4$ in \cref{fig:crystalgfn}). 
Every space group is classified into a lattice system which, in turn, constrains the values of the lattice parameters. 
For instance, for the cubic lattice system, $a=b=c$ and $\alpha=\beta=\gamma=90^{\circ}$. 
The remaining constraints are illustrated in \cref{fig:latticesystems}.
This drastically reduces the state space and removes the need for having to learn such relationships and constraints from the data.

\section{Target properties / Reward functions}
\label{sec:reward-function}

As explained in~\cref{sec:background}, Crystal-GFN is trained to sample proportionally to a reward function, offering the flexibility to use as reward any quantity of interest, like a desired material property. In this work, we focus on 3 properties: unit cell's density, formation energy (FE) per atom and the band gap (BG) or energy gap. 
From now on, \textit{Crystal-GFN} will refer to the generic method, while specific instances trained for different target properties will be called \textit{Crystal-GFN (De)}, \textit{Crystal-GFN (FE)} and \textit{Crystal-GFN (BG)}, respectively for the density, formation energy and band gap targets.

\subsection{Density}
\label{sec:density}

The density of a crystal's unit cell is defined as the total mass of the atoms divided by the volume of the cell. 
We can compute the total mass from the composition, since we know the atomic mass of each element and the number of atoms per element. 
Similarly, we can also compute the volume from the lattice parameters. 
The density is a convenient property for a proof of concept because we can calculate it exactly, and also because we know what crystal properties maximise it. 
Namely, we expect Crystal-GFN (De) to tend to sample crystals that have low lattice lengths $(a, b, c)$, highly packed space groups, large numbers of atoms, and elements of high atomic mass.

\subsection{Formation energy}
\label{sec:formationenergy}

The formation energy of a material is defined as the amount of energy needed to form a material from its constituent elements in their standard states. 
If a compound $C$ is the combination of elements $A$ and $B$, that is $A + B \rightarrow C$, then the formation energy ($E_f$) of $C$ is defined as follows: 
$E_f(C) = E(C) - E(A) - E(B)$.
A negative formation energy indicates that the formation reaction of a material from its constituent elements is spontaneous or favourable. 
On the contrary, a positive formation energy indicates that the formation reaction is not thermodynamically favoured.

Because crystal structures sampled by Crystal-GFN (FE) may be unknown or not characterised in existing databases, their true FE may be unknown too. Accurate determination of FE using Density Functional Theory \citep{kohn1996density} is computationally intractable for large-scale exploration.
This is why we design and train a \textit{proxy} machine learning model to predict the FE given a crystal $x \in \mathcal{X}$, which is parameterised as the output of Crystal-GFN (FE). See \cref{sec:dave} for further details about the model.
The final reward function is a Boltzmann transformation of the proxy output, $h_{FE}(x)$: $R(x) = \exp\big(-\frac{h_{FE}(x)}{T}\big)$ where $T$ is a temperature hyper-parameter.
This ensures that lower FE yields higher positive reward, and we can control our preference for lower energies with the temperature $T$.

\subsection{Band gap}
\label{sec:bandgap}

Similar to FE, we demonstrate our method's flexibility by training a Crystal-GFN (BG) to sample materials with a target electronic band gap, defined as the difference between the lowest unoccupied conduction band and the highest occupied valence band. In other words, it is the minimal energy difference an electron needs to be in a conduction band. Being able to generate materials with specific band gaps is useful, for instance, for photovoltaics. In our experiments, we choose 1.34 eV as our target, which corresponds to the theoretical optimum for a p-n solar cell, known as the Shockley–Queisser limit \citep{shockleyDetailedBalanceLimit1961a}.
Given a predicted band gap $h_{BG}(x)$, we use $R(x) = \exp(-\frac{(h_{BG}(x) - 1.34)^2}{T})$ as reward signal.

\subsection{Proxy ML models of FE and BG}
\label{sec:dave}

To predict a sample's formation energy or band gap, we train a physics-informed Multi-Layer Perceptron (MLP) for each property on the MatBench data set~\citep{dunn2020benchmarking}, a snapshot of the Materials Project \citep{2013materialsproject}.
For a given crystal structure, the input to the proxy MLP is the concatenation of: 1) a physical embedding of the crystal's elements, obtained using PhAST~\citep{duval2022phast} by encoding the atomic number, the period, the group and other relevant atomic properties; 2) a learned embedding for its space group ; 3) the standardised lattice parameters. See \Cref{sec:app:proxy} for more details.
We stratify the MatBench data set into train, validation and test sets, controlling for the distribution in each target property. 
Our training and validation sets contain 65,048 and 23,232 data points, respectively.



\section{Empirical evaluation}
\label{sec:empirical-evaluation}

In this section, we present the results of experiments designed to evaluate the potential of Crystal-GFN to explore the space of materials.
The motivation of using GFlowNet to generate crystal structures is twofold. 
One, we are interested in discovering structures with high scores of a property of interest, in our cases: high density, low formation energy or band gap close to 1.34 eV.
At the same time, we want to discover not just one but multiple and diverse structures.
The main reason to seek diverse samples is that often the target function is underspecified and the true objective is multifaceted or unknown.
A natural way of dealing with underspecification is to try multiple candidates to increase the likelihood of finding successful structures for the downstream applications \cite{jain2023scientificdiscovery,hg2023multifidelity}.
Thus, we here evaluate two aspects of Crystal-GFN: the distribution of the predicted property of interest for the generated samples, and sample diversity.

\subsection{Experimental setup}
\label{sec:experimental-setup}

Exploring the infinitely large space of crystals to find the ones with certain properties is a daunting ``needle-in-a-haystack'' task.
In order to make the haystack a little smaller---though still infinitely large---we restrict the search task to a subset of 113 space groups and compositions with up to 5 unique elements from a set of 22 common elements, and up to 16 atoms per element, up to 80 atoms in total. 
We note that this still a much larger search space than many other works in the literature, that are often restricted to specific lattices \citep{zhao2021high}, ternary compounds \citep{zhu2023wycryst} or a smaller number of atoms \citep{zeni2023mattergen}.

We train all Crystal-GFN models for 50,000 iterations, which amounts to 500,000 queries to the proxy model and under 30 hours on a CPU-only machine.
Further details about the experimental setup are provided in~\cref{sup:experimental-setup}.

\subsection{Density}
\label{sec:results-density}

As the first proof-of-concept experiment, we trained Crystal-GFN to sample high-density crystals. We observe that our model learned to sample low lattice lengths (a,b,c) with the mean value $4.9~\si{\angstrom}$ compared to $8.3~\si{\angstrom}$ in the validation dataset. Additionally, we clearly see a shift towards spacegroups from the cubic, tetragonal, and hexagonal family (81, 99, 115, 195, 200, 207, 215, 221). These crystal systems indeed tend to have a more compact packing. Element occurrence  distribution in the 100 most dense samples is diverse and clearly shifted towards heavier atoms (see Fig. \ref{fig:app:top-dense-elems}). Finally, the average number of atoms in the 100 most dense samples is $58.8$ compared to $48.1$ over the validation set.

\subsection{Formation energy}
\label{sec:results-fe}

In this section, we present the results from the experiments to generate materials with low formation energy. We discuss the performance of our proxy model, the distribution of FE in Crystal-GFN samples and the diversity of these samples.

\paragraph{Proxy model of the formation energy}

As presented in~\cref{sec:dave}, our FE reward function is based on a proxy MLP trained on MatBench (Materials Project).
Since part of our performance analysis for the proposed Crystal-GFN depends on the formation energy prediction  by this proxy model, it is important to first verify its accuracy.
In contrast to other methods, it does not rely on atom positions to predict the FE of a crystal, but rather on the crystal higher-level description, namely its space group, composition and lattice parameters, akin to the outputs of Crystal-GFN (FE).
Nonetheless, it achieves a mean absolute error (MAE) of $0.10 \pm 0.005\ \text{eV/atom}$ on the validation set, within the theoretical accuracy of DFT which is 0.08 - 0.1 eV~\citep{Bogojeskina2020quantum}. It also ranks 15th on the Matbench Leaderboard for this specific task \cite{dunn2020benchmarking}.
We further describe its hyper-parameters and detailed performance in \cref{sec:app:proxy}.

\paragraph{Predicted formation energy of samples}

\begin{figure}[htb]
    \centering
    \includegraphics[width=\columnwidth]{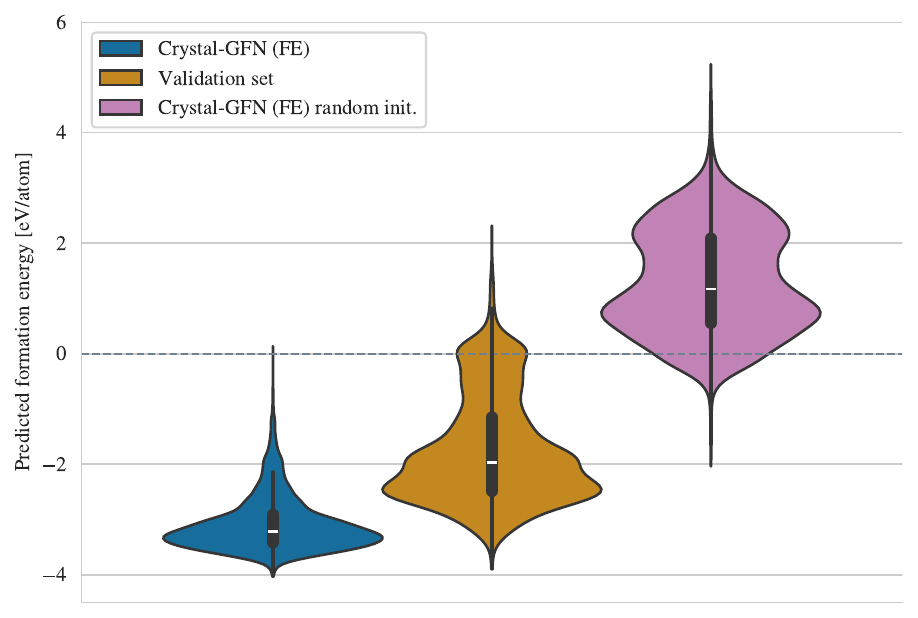}
    \caption{Distributions of the formation energy predicted by our proxy model in three relevant distributions of samples: from Crystal-GFN (FE) after training, from the validation set and from an untrained Crystal-GFN (FE). As a main conclusion, we observe that Crystal-GFN (FE), after training, manages to sample crystals with predicted formation energies considerably lower than the bulk of the validation set. Note that the shift in the distribution is significantly larger than the error of the predictive model.}
    \label{fig:distributions}
\end{figure}

Evaluating generative models is well known to be a difficult, task-dependent problem.
Here, in order to gain insights about the sampling policy of the trained Crystal-GFN (FE), we sample 10,000 crystals and compare the distribution with 1) the validation set from MatBench and 2) 10,000 samples from a randomly initialised and untrained Crystal-GFN (FE).
\Cref{fig:distributions} shows the kernel density estimation of the formation energies predicted by our proxy model in the aforementioned sets of samples.
As main conclusion, we observe that training the Crystal-GFN shifts the formation energy of the sampled crystals towards lower values, with a median of -3.2 eV/atom, while the median energy in the validation set is -2 eV/atom. 
Given the vast search space, it is remarkable that with only 30 hours of training on 1 CPU and 500,000 queries to the proxy model, Crystal-GFN (FE) learns to sample from a distribution where 95~\% of the structures have a formation energy lower than -2 eV/atom (as predicted by the proxy model). Moreover, as shown in Figure \ref{fig:ehull_distributions}, 65.6\% of the materials are predicted to lie below the convex hull of the Materials Project.

\paragraph{Diversity of samples}

As discussed before, an important goal in certain materials discovery applications is to find \textit{diverse} candidates. 
In other words, sampling crystals with low formation energy would not be useful for most applications if all the crystals were identical or highly similar. 
To gain insights about the diversity, we analyse the crystal structures sampled by the trained Crystal-GFN (FE). 
The main conclusion is that it samples candidates with very high diversity. 
In particular, regarding the composition, we find that all 22 elements are found in the samples from the Crystal-GFN (FE), and 15 of them appear in the 100 samples with lowest predicted formation energy. 
Regarding space groups, we find that 4 out of the 8 crystal-lattice systems and 4 of the 5 point symmetries are found in the top-100 samples alone. 
While not all 113 space groups are found in the 10,000 samples, 73 of them were present (65~\%). 
Finally, in terms of lattice parameters, we also observe relatively similar distributions of lengths and angles, compared to the MatBench data set. 
Further details and visualisations of these results are provided in \cref{sup:results}. We additionally provide a study of the samples' validity in~\Cref{sec:app:validity}.

\subsection{Band gap}
\label{sec:results:bg}

Similarly to the formation energy results, we describe here the results for our Crystal-GFN (BG) trained to sample diverse crystal structures with a band gap close to 1.34 eV.

\paragraph{Proxy model of the band gap}
As per the previous sections, we train a proxy MLP with a similar architecture (though different hyper-parameters) as the one described for the formation energy. After 150 epochs, our band gap proxy model achieves a performance of 0.321 $\pm$ 0.03 eV.

\begin{figure}[htb]
    \centering
    \includegraphics[width=\columnwidth]{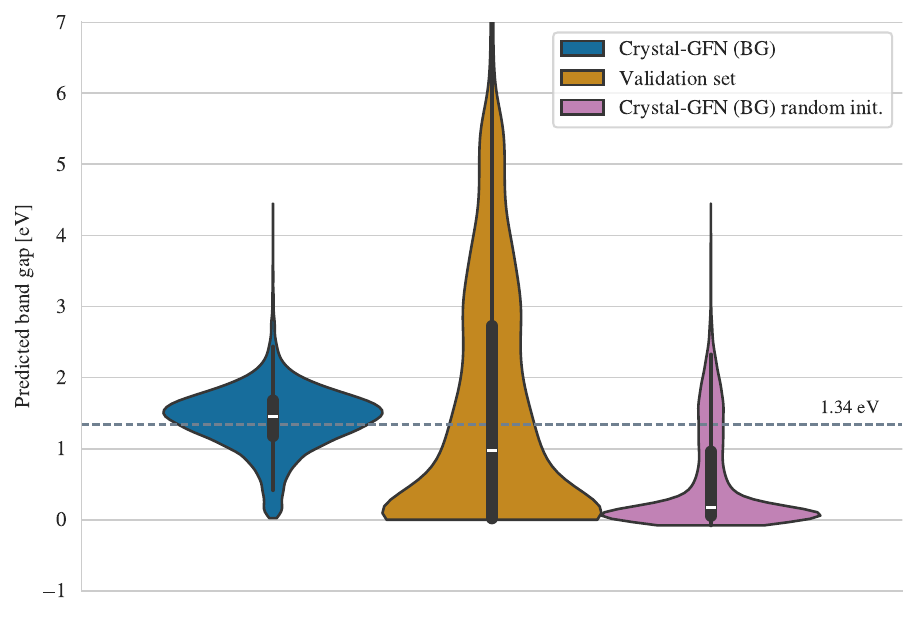}
    \caption{Distributions of the band gap predicted by our proxy model in three relevant distributions, similarly to~\Cref{fig:distributions}. We observe that the trained Crystal-GFN (BG) is able to sample crystals with predicted band gaps close to the target value of 1.34 eV.}
    \label{fig:bg-distributions}
\end{figure}

\paragraph{Predicted band gap of samples} We present in~\Cref{fig:bg-distributions} the distribution of the band gap (as predicted by the proxy) for 10k samples from a trained Crystal-GFN (BG), an untrained Crystal-GFN (BG) and the validation set. We observe that the trained model is able to sample structures often close to the target band gap of 1.34 eV. Note that Crystal-GFN (BG) is not trained to sample \textit{only} structures with that band gap, but \textit{proportionally} to the distance. This explains the peak close, but not restricted, to the target.

\subsection{Restricted sampling}
\label{sec:conditional}

Certain applications often require particular classes of materials, for example with specific elements or symmetries. It may also be convenient to restrict the number of atoms in the unit cell to speed up DFT simulations. The flexibility of Crystal-GFN allows us to easily restrict the sampling space of a pre-trained model.
Here, we use the Crystal-GFN (FE) described in \cref{sec:results-fe}, with the experimental setup detailed in \cref{sec:experimental-setup} and we sample 1,000 samples under four different restrictions that cover the three subspaces---space group, composition and lattice parameters: 

\begin{itemize}
    \item A: The composition is restricted to only elements Fe and O, with a maximum of 10 atoms per element.
    \item B: We sample in the ternary space for Li-Mn-O, keeping the element count to maximum 16 atoms.
    \item C: We restrict the space groups to only cubic lattices.
    \item D: We restrict the range of the lattice parameters to lengths between 10 and 20 angstroms and angles between 75 and 135 degrees.
\end{itemize}

\begin{figure}[h]
    \centering
    \includegraphics[width=\columnwidth]{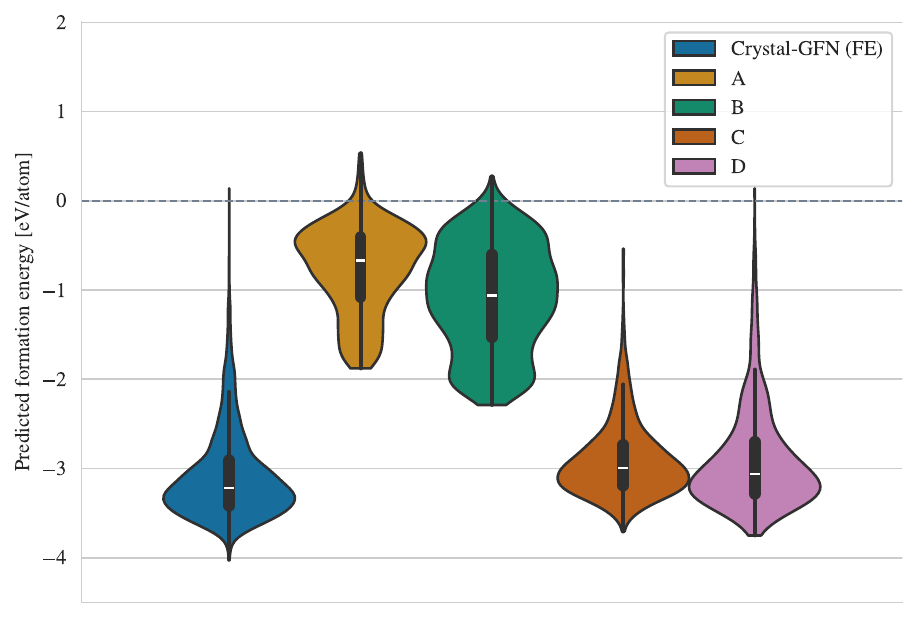}
    \caption{Distributions of the predicted formation energy in the sets of 1,000 samples generated under each restriction (A--D), plotted alongside the original set sampled without constraints. We observe that Crystal-GFN (FE) is not only able to sample in a restricted space, but it also keeps the FE low.}
    \label{fig:conditionalsampling}
\end{figure}

In \cref{fig:conditionalsampling}, we observe that the formation energies of the samples with restrictions remain low, even though the model was trained without the restrictions.

\section{Discussion, conclusions and future work}
\label{sec:conclusions}

In this paper, we have introduced a new generative model named Crystal-GFN for sampling inorganic crystals proportionally to a property of interest. Through an empirical evaluation, we have demonstrated several distinctive and desirable features of Crystal-GFN: 

\begin{itemize}
\item Thanks to the sequential construction of crystals in a space inspired by theoretical crystallography, Crystal-GFN is able to flexibly incorporate constraints regarding the charge neutrality of the composition, the compatibility between composition and space group, as well as between space group and lattice parameters. Additional constraints could be easily introduced in the future.
\item In our experiments, we have shown that Crystal-GFN can be flexibly trained with any available reward function. We have sampled materials with low predicted formation energy, a predicted band gap close to a targeted value and high density. 
\item Crystal-GFN excels at sampling highly diverse crystals, in terms of coverage of space groups, compositions and lattice parameters.
\item Crystal-GFN is also highly flexible regarding the definition of the sampling space. The set of elements, space groups, number of atoms, range of the lattice parameters, can all be adjusted before training a model. Remarkably, the sampling space can be further constrained at sampling time to meet specific requirements.
\item Finally, unlike generative models based on diffusion, both training and sampling are notably fast. Training was carried out in under 30 hours in a CPU-only machine and we generated each set of 10,000 samples in a few minutes.
\end{itemize}

Interesting directions for future work include extending Crystal-GFN with more domain-inspired constraints as well as with additional subspaces to sample the atomic positions in the unit cell.
Alternatively, and as stated previously, coordinates
can be retrieved by relaxation, using conventional or machine learning interatomic potentials.
Furthermore, it would be interesting to explore other properties of interest beyond the formation energy of the sampled crystals.

A link to the code repository with the instructions for the reproduction of the results will be included in the manuscript upon acceptance.

\section*{Impact Statement} Our work is motivated by the climate crisis, the need to develop sustainable technology and the ambition to apply our approach to materials discovery. However, as with other related work, there is a potential risk of dual use of the technology by nefarious actors~\citep{urbina2022dual}. The authors strongly oppose any uses or derivations of this work intended to cause harm to humans or the environment. 

\bibliography{references}
\bibliographystyle{latex-simple}

\newpage
\appendix
\onecolumn
\section{Materials and crystal structures}
\label{sec:app:crystals}

\paragraph{Materials} Materials encompass the substances that make up objects. Their distinctive characteristics arise from a combination of factors, including their composition, structure, and manufacturing processes. While materials come in various forms, our primary focus will be on solid-state materials, specifically crystals, owing to their significance in many machine learning applications we have encountered.

\paragraph{Solid-state materials} Solid-state materials consist of a multitude of smaller building blocks, such as atoms or molecules, securely positioned in fixed locations. They exhibit high-density packing and strong mutual attraction, resulting in a stable structure with a well-defined volume. Materials that possess a well-ordered, infinitely repetitive structure are categorised as crystals, while those lacking such long-range positional order are termed amorphous.

\paragraph{Crystals} Also known as crystalline materials, crystals represent a subset of solid materials characterised by their periodic structure. This consists in a repeated and coherent placement of atoms along all spatial dimensions. It is this inherent regularity that sets crystals apart and confers them unique properties. Crystals manifest in diverse forms and showcase exceptional attributes like magnetism, transparency and/or elevated melting points. These functional properties offer many opportunities to fulfill a wide range of needs.


Mathematically, a crystal can be defined as a basis motif, called the unit cell, repeated infinitely in the 3 directions of space, along a lattice. The lattice is defined by three lattice basis vectors $\mathbf{a}_{i\in (1,2,3)}$, that can be represented by six parameters: their length $a, b, c$ and the angles between them, $\alpha$, $\beta$, $\gamma$. All the atoms in the infinite crystal can therefore be reduced to their finite set $S$ in the unit cell. These are typically expressed in the basis of the lattice vectors: $S = \cbrace{\pr{Z_i, \mathbf{x}_i} \mid 0 \leq x_{i,j\in(1,2,3)} <1 }$, where $\mathbf{x}_i$ are the reduced coordinates and $Z_i$ their atomic number. The crystal is therefore defined by all atoms $\cbrace{\pr{Z_i, (h+x_{i1})\mathbf{a}_1+(k+x_{i2})\mathbf{a}_2+(l+x_{i3})\mathbf{a}_3} \mid h,k,l \in \mathbb{Z} }$ where $i $ runs over all sites.

Different unit cells are possible, and of particular prominence is the primitive cell $U$, which consists of the \emph{smallest} unit cell that still contains all information of the crystal.

\paragraph{Periodic boundary conditions} Periodic boundary conditions (PBC) are a computational strategy widely utilized to model the infinitely repeating lattice structure. Instead of limiting the simulation to a single, finite unit cell that contains a fixed number of atoms or molecules, PBC assume an infinite tessellation of identical unit cells extending in all spatial directions. Consequently, when an atom or molecule crosses the boundary of one cell, it is seamlessly reintegrated into the system from the corresponding opposite boundary, mirroring its continuous motion through an unbounded lattice. This method allows for the realistic simulation of bulk material properties and behaviors within a computationally manageable domain.

\paragraph{Wyckoff positions} Wyckoff positions serve as a systematic method for categorizing the locations of atoms within a crystal lattice based on symmetry considerations. Defined as classes of crystallographic orbits, these positions limits the coordinates an atom can occupy to respect the lattice's symmetry elements (e.g., rotations, reflections, inversions). The concept of Wyckoff multiplicity refers to the number of equivalent locations that an atom occupies as a result of these symmetry operations.

\section{Experimental setup}
\label{sup:experimental-setup}

In this section, we provide additional details about the hyper-parameters and experimental setup used to train the Crystal-GFN to obtain the results presented in \cref{sec:empirical-evaluation}.

In order to reduce the search space in our experiments, we apply the following restrictions:

\begin{itemize}
    \item Compositions consist of up to 5 different elements from the subset of these 22 elements: H, Li, B, C, N, O, F, Na, Mg, Al, Si, P, S, Cl, K, V, Mn, Fe, Co, Ni, Cu, Se. These are the most common elements in the MatBench data set used to train the proxy model. Note that even this reduced set yields a combinatorially large search space.
    \item Compositions can contain up to 16 atoms per element (80 atoms in total). This number is obtained by finding the lowest Wyckoff position multiplicity for each space group, and then computing the maximum of these values across all space groups. This means that 16 is the lowest possible value that still makes it possible for the Crystal-GFN to generate samples from all space groups while respecting their symmetry constraints.
    \item Space groups are the intersection of train and validation space groups, from the proxy model's data set. There are 113 of them.
    \item The minimum and maximum lengths of the unit cell are 0.9 and 100 angstroms, respectively; the minimum and maximum angles are $50^{\circ}$ and $150^{\circ}$, respectively. These values are set so as to include the bulk of the training and validation sets, excluding outliers.
\end{itemize}

We train the Crystal-GFN by sampling 10 trajectories per iteration from the current forward policy. In order to encourage further exploration, 10~\% of the steps in the trajectories are sampled at random from a uniform distribution. In total, we train for 50,000 iterations, which amounts to 500,000 queries to the proxy model. This took about 12 hours on a CPU-only machine.

The architecture of both the forward and the backward GFlowNet policy models is a 3-layer multi-layer perceptron with 256 units per layer. We trained with the Adam optimiser and a learning rate of 0.0001. As is common with the Trajectory Balance objective \citep{malkin2022tb}, we set a higher learning rate (0.01) for the 16 learning weights used to parameterise the partition function. In order to sample structures with lower (negative) formation energies, we set a temperature $T=8$ in the reward function defined in \cref{sec:reward-function}.

The distribution to sample the increments of the lattice parameters subspace is a mixture of 5 Beta distributions. In order to ensure numerical stability during training, we restrict the values of the coefficients of the Beta distributions to the range $[0.1, 100]$. One of the conditions that must be satisfied by generalised GFlowNets is that trajectories must have a finite length. To this end, we set the minimum increment to 10~\% of the range of each dimension.

\section{Validity of Samples}
\label{sec:app:validity}

\paragraph{Charge neutrality and electronegativity} While charge neutrality is not the end-all be-all of valid crystals, it is an important principle that domain experts keep in mind when designing new materials. Similarly, electronegativity can also be used to implement an empirical test: the most electronegative element should possess the most negative charge (known as the Pauling rule). We use the Python library SMACT \cite{Davies2019} to screen generated samples and calculate that 84~\% of the Crystal-GFN (FE) samples pass the aforementioned criteria. 
We note that we do not achieve 100 \% because we trained with a slightly different set of oxidation states to implement constraint $C_3$ than SMACT's, and that the Pauling rule was not explicitly constrained. We leave this improvement for future work.

\paragraph{Compatibility between composition and space group} The elements and their counts in crystals should be compatible with the multiplicities of the Wyckoff positions available in the selected space group. We use Python library Pyxtal \cite{Fredericks2021} to perform this structural check and achieve a 100 \% success rate, given our hard constraint $C_2$.

\paragraph{Crystal system and lattice parameters} The crystal lattice systems define the relations between the 6 lattice parameters. Our samples pass this check easily thanks to constraint $C_4$, achieving 100 \% success rate as compared to an abysmal 0.41 \% for a GFN trained without constraints.

\section{Proxy MLP}
\label{sec:app:proxy}
\subsection{Architecture}

For a given crystal structure, the input to the proxy MLP is the concatenation of: a physical embedding of the crystal's elements using PhAST~\citep{duval2022phast}, a learned embedding for its space group and the standardised lattice parameters. We stratify the MatBench data set into train, validation and test sets, matching their distributions of target FE. The final reward function is a Boltzmann version of the proxy: $R(x) = \exp\big(-\frac{\text{MLP}(x)}{T}\big)$ where $T$ is a temperature hyper-parameter. This ensures that a lower FE yields higher positive reward, and we can control our preference for lower energies with the temperature $T$.

Overall, the architecture consists of concatenated representations for the composition ($h_C$), the space group ($h_{SG}$) and the lattice parameters ($h_{LP}$) that are used as input to a prediction MLP $\hat{y} = \text{MLP}_{out}([h_C, h_{SG}, h_{LP}])$. In the following we describe how each $h$ is obtained from the data:

\begin{itemize}
\item $\mathbf{h_C}$: Each element $Z_i$ in the composition is embedded with PhAST~\citep{duval2022phast}: it is a concatenation of 1/ static physical properties $P_Z$ projected into some latent space with a small MLP 2/ a learned embedding for the atomic number $Z$, Period $P$ and the Group $G$ of the element, noted $E_i$: 

\begin{gather}
    h_C = \text{MLP}_C\Big(\big[
            \text{MLP}_P(P_Z), E_Z, E_P, E_G  
            \big]\Big)
\end{gather}

\item $\mathbf{h_{SG}}$: Each space group is embedded as a lookup in a learnable embedding matrix.

\item $\mathbf{h_{LP}}$: Lattice parameters are first standardised element-wise using the training set statistics. This 6-dimensional vector is then passed through a small MLP:

\begin{gather}
    h_{LP} = \text{MLP}_{LP}\bigg(\frac{\text{LP}(x) - \mu_{train}(\text{LP})}{\sigma_{train}(\text{LP})}\bigg)
\end{gather}

\end{itemize}

\subsection{Performance}

\subsubsection{Formation Energy}
\label{sec:app:proxy:perf}
\begin{figure}
    \centering
    \includegraphics[width=0.75\textwidth]{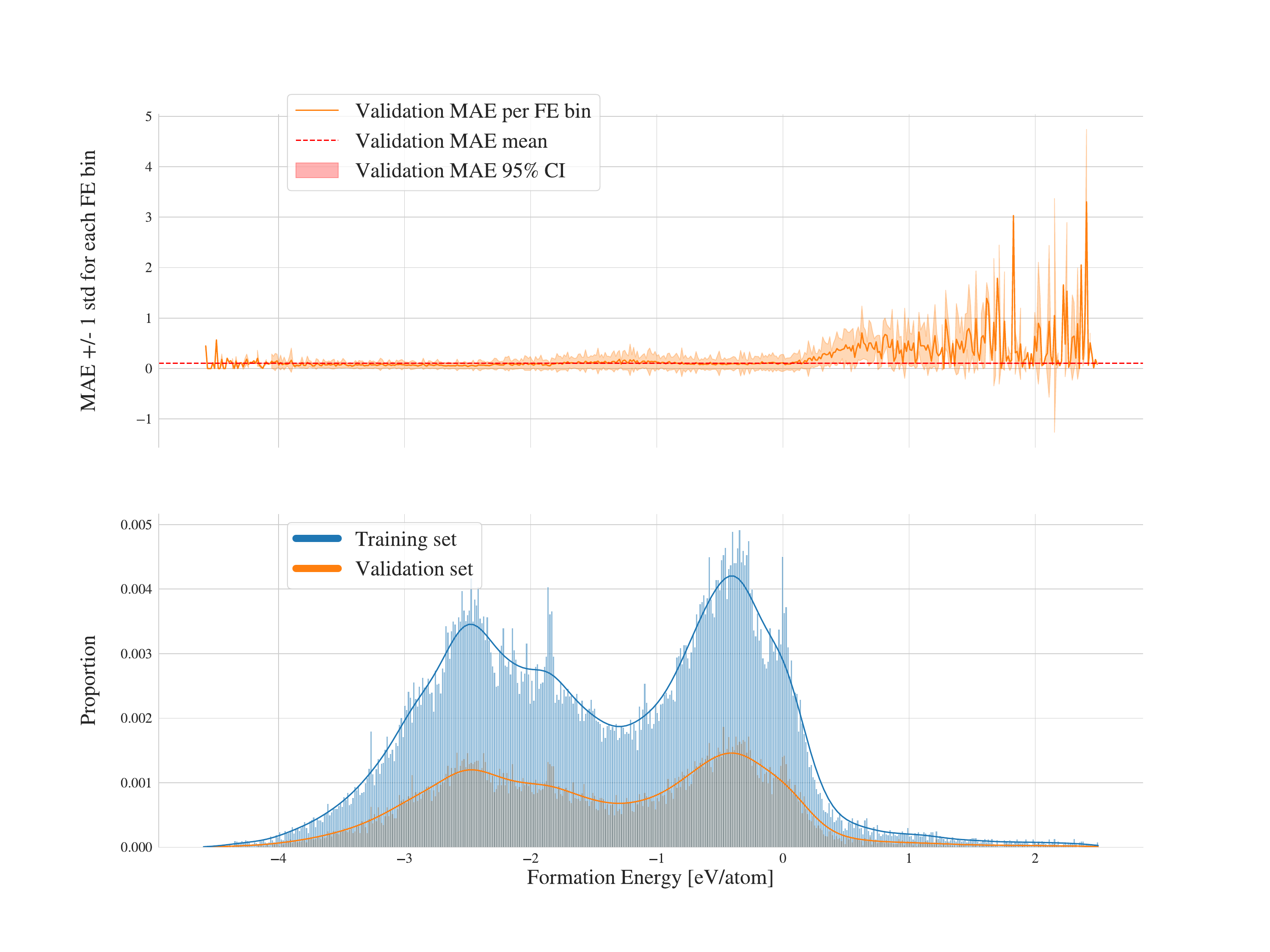}
    \caption{Formation Energy (FE) proxy MLP performance on the validation set and data split FE distributions. The average MAE is $0.10~\text{eV/atom}$. We can also see the effect of the stratification algorithm which yields similar FE distributions between the train and validation data set splits.}
    \label{fig:proxy-mae-std}
    \vspace{1cm}
    \centering
    \includegraphics[width=0.75\textwidth]{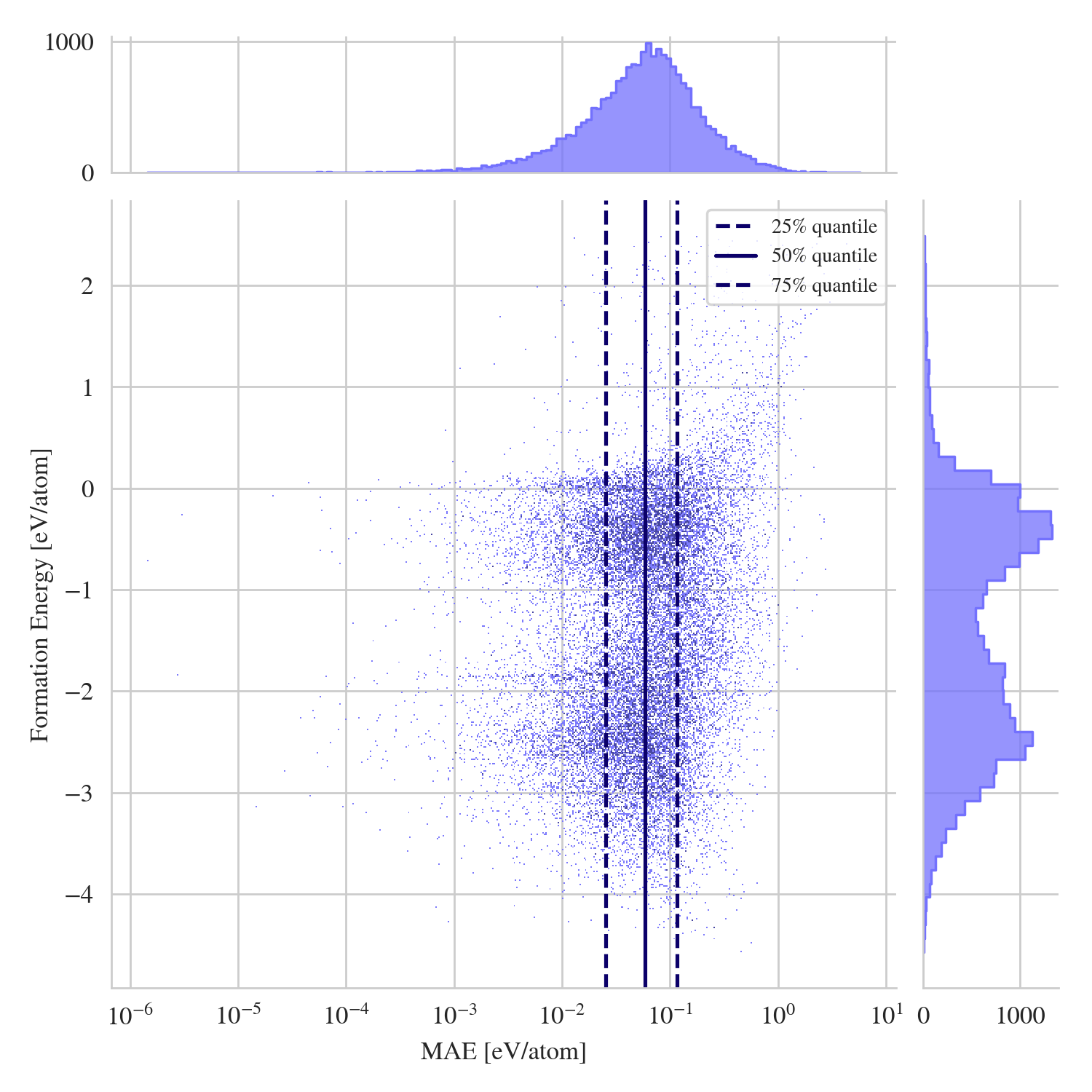}
    \caption{Distribution of FE values in the validation set and associated FE proxy MLP MAE, with $25~\%, 50~\%$ and $75~\%$ MAE quantiles.}
    \label{fig:proxy-mae-joint-plots}
\end{figure}

Our FE proxy MLP model achieves an overall Mean Absolute Error of $0.10 \pm 0.005\ \text{eV/atom}$\footnote{95\% confidence interval, computed modeling the MAE as a log-normal distribution. This is further justified by~\cref{fig:proxy-mae-joint-plots}.}. For reference, best performing methods in MatBench leaderboard\footnote{\url{https://matbench.materialsproject.org/Leaderboards\%20Per-Task/matbench_v0.1_matbench_mp_e_form/}} can achieve up to 10$\times$ better performance ($0.0170$ eV/atom). However, they all use 3D positions of the crystals, and remarkably, our approach outperforms some of them, indicating that leveraging composition, space-group and lattice parameters serves as a viable representation in the formation energy estimation task.

In~\Cref{fig:proxy-mae-std,fig:proxy-mae-joint-plots} we detail further the performance of our FE proxy MLP on the MatBench validation set. In particular we observe that it maintains good performance for structures with FE up to $\sim0.3~ \text{eV/atom}$. The MAE beyond that FE level becomes high and unstable, but this is expected as those are the regions of the target space with the least data available. Additionally we can verify in~\Cref{fig:proxy-mae-std} that our stratification algorithm works as expected, with similar FE distributions in the validation and train splits.

\subsubsection{Band Gap}

Our BG proxy MLP model achieves an overall Mean Absolute Error of $0.32 \pm 0.033\ \text{eV}$. For reference, best performing methods in MatBench leaderboard\footnote{\url{https://matbench.materialsproject.org/Leaderboards\%20Per-Task/matbench_v0.1_matbench_mp_gap/}} can achieve up to 2$\times$ better performance ($0.1559$ eV). However, they all use 3D positions of the crystals, and remarkably, our approach outperforms some of them, indicating that leveraging composition, space-group and lattice parameters serves as a viable representation in the band gap estimation task.

\begin{figure}[]
    \centering
    \includegraphics[width=0.75\textwidth]{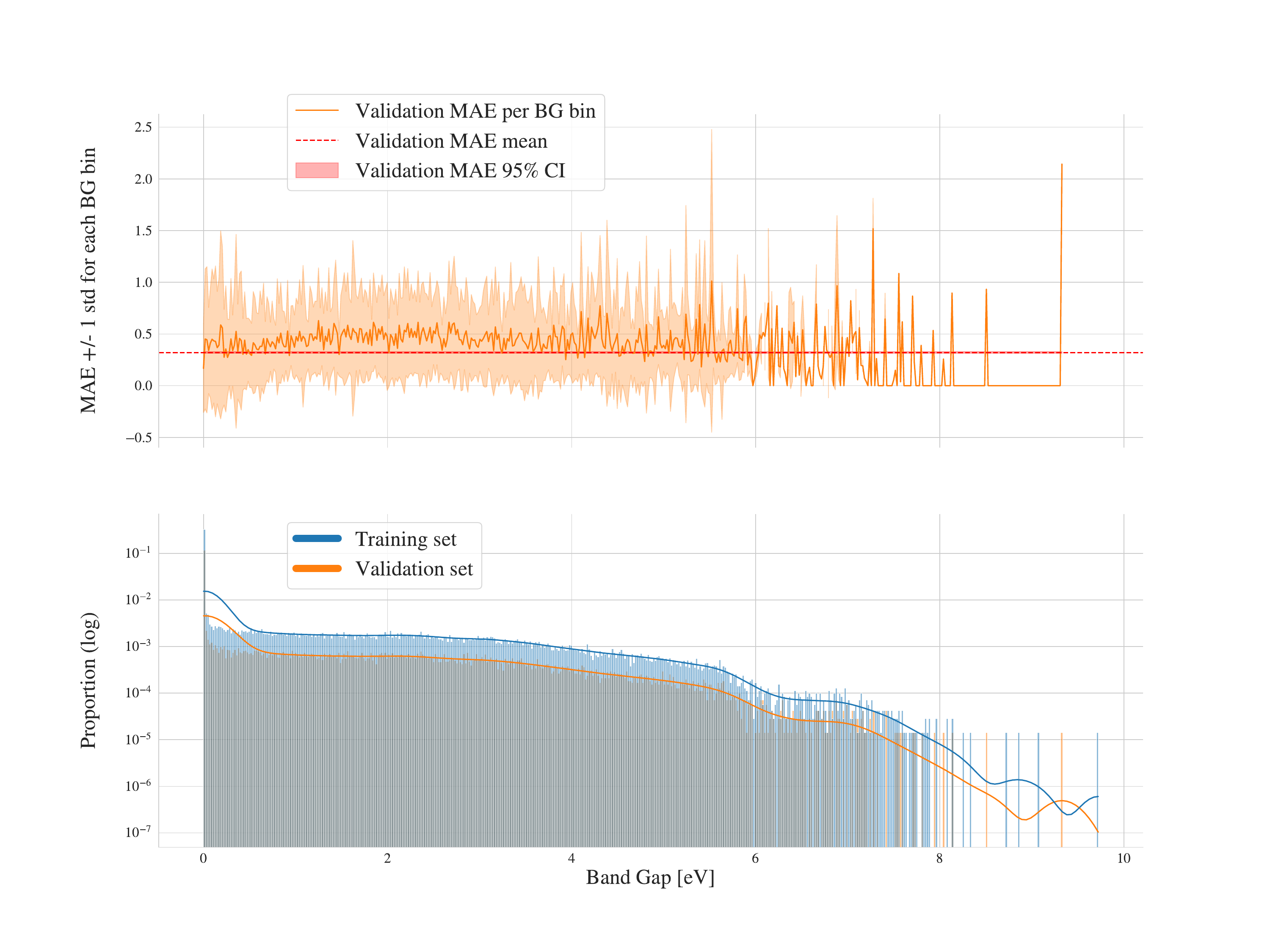}
    \caption{Band Gap (BG) proxy MLP performance on the validation set and data split band gap distributions. The average MAE is $0.32~\text{eV}$. We can also see the effect of the stratification algorithm which yields similar BG distributions between the train and validation data set splits.}
    \label{fig:bg-proxy-mae-std}
    \vspace{1cm}
    \includegraphics[width=0.75\textwidth]{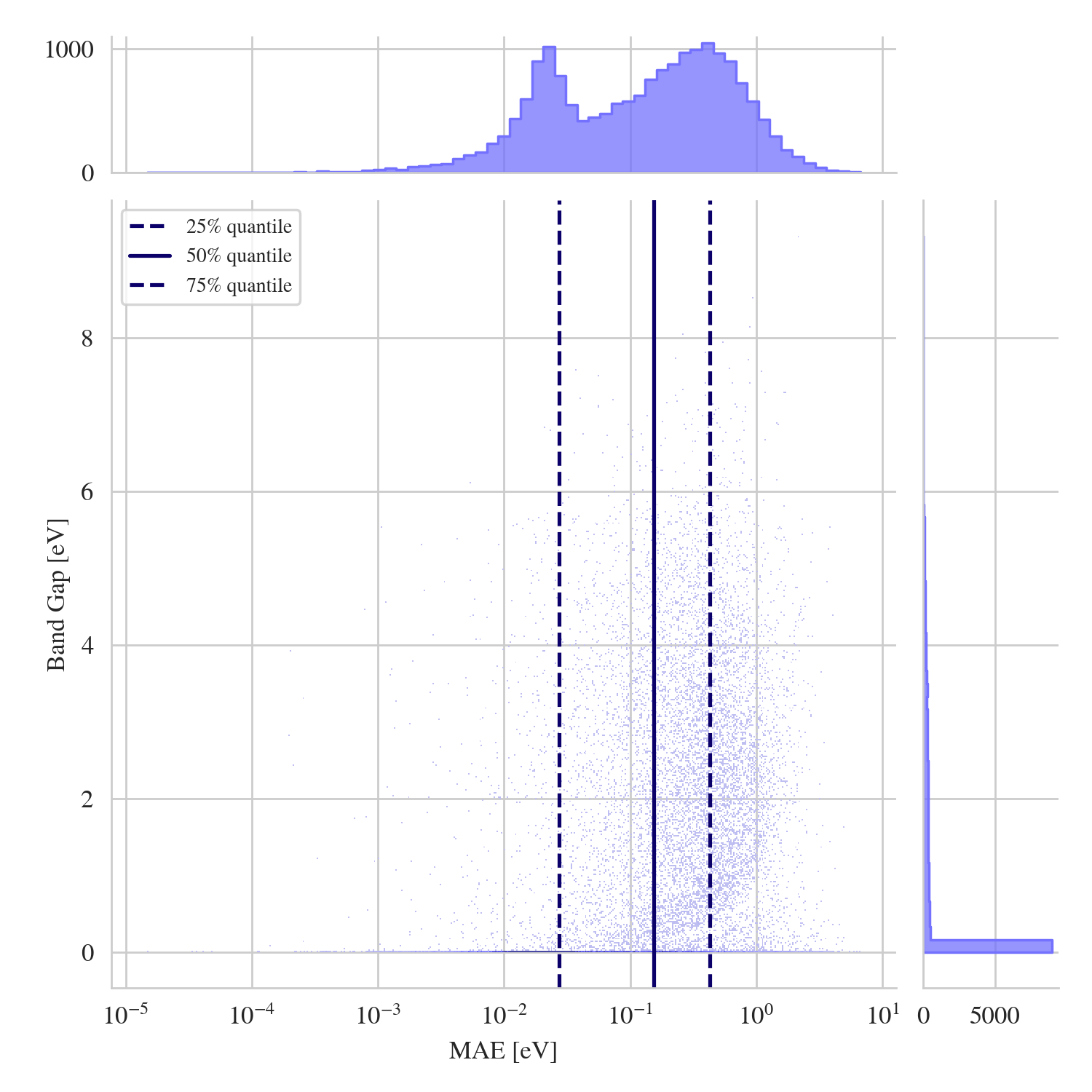}
    \caption{Distribution of band gap values in the validation set and associated BG proxy MLP MAE, with $25~\%, 50~\%$ and $75~\%$ MAE quantiles.}
    \label{fig:bg-proxy-mae-joint-plots}
\end{figure}

In~\Cref{fig:bg-proxy-mae-std,fig:bg-proxy-mae-joint-plots} we detail further the performance of our BG proxy MLP on the MatBench validation set. In particular we observe that it a rather homogeneous performance across band gap values.

\subsection{Hyper-parameters}
\label{sec:app:proxy:hp}

We present the hyper-parameters of our two Proxy MLPs (FE and BG) in~\Cref{tab:proxy-hp} along with a description of their role in the architectures.

\begin{table}[]
\centering
\begin{adjustbox}{width=\textwidth}
\begin{tabular}{lccl}
\textbf{Hyper parameter} & \textbf{FE} & \textbf{BG} & \textbf{Description} \\ \hline
\texttt{properties\_proj\_size}     & 64      &   384      & Projection size of atomic properties $h = \mathbf{W}p_Z$ \\
\texttt{group\_emb\_size}           & 16      &   160      & Embedding size of element's group                            \\
\texttt{period\_emb\_size}          & 256     &   160      & Embedding size of element's period                           \\
\texttt{z\_emb\_size}               & 128     &   384      & Embedding size of element's atomic number                    \\
\texttt{sg\_emb\_size}              & 128     &   416      & Crystal space group embedding size                           \\
\texttt{lat\_hidden\_channels}      & 284     &   448      & Hidden channels for layers processing the lattice parameters \\
\texttt{lat\_num\_layers}           & 1       &     3      & Number of hidden layers for the lattice parameters MLP       \\
\texttt{num\_layers}                & 5       &     7      & \begin{tabular}[c]{@{}l@{}}Number of layers for final MLP processing composition, \\ space group and lattice parameters hidden representations\end{tabular} \\
\texttt{hidden\_channels}           & 576     &   1856     & Size of final MLP hidden layers.                             \\
\texttt{optimiser}                  & Adam    &   Adam     & Optimiser                                                    \\
\texttt{lr}                         & 0.0017  &   0.0018    & Learning rate                                                \\
\texttt{batch\_size}                & 448     &   192      & Batch size                                                   \\
\texttt{scheduler}                  & ReduceLROnPlateau & ReduceLROnPlateau & \begin{tabular}[c]{@{}l@{}}Divide learning rate by 2 when validation MAE \\ does not improve for 4 epochs\end{tabular} \\
\texttt{es\_patience}               & 11      &   9    & Early stopping patience (in epochs)                          \\
\texttt{epochs}                     & 100     &   150    & Max epochs before early stopping                             \\
\end{tabular}
\end{adjustbox}
\vspace{0.5cm}
\caption{Hyper-parameters of the Formation Energy and Band Gap MLP proxies and their values used in our models.}
\label{tab:proxy-hp}
\end{table}

\section{Additional results}
\label{sup:results}

In this section, we provide additional results for further evaluation of our proposed Crystal-GFN.

\subsection{Predicted energy above hull}

The energy above hull is computed using the pymatgen PhaseDiagram. All compositions and their corresponding predicted formation energies are compared to the convex hull formed by a snapshot of the Materials Project (as of March 2021). A compound is defined stable if it has a negative energy above hull (below 0 eV/atom), thus effectively belonging below the simplex. Computing the hull takes around 48 CPU hours, while computing the distance for any new compound takes around 300 ms.

\begin{figure}[h]
    \centering
    \includegraphics[width=0.7\columnwidth]{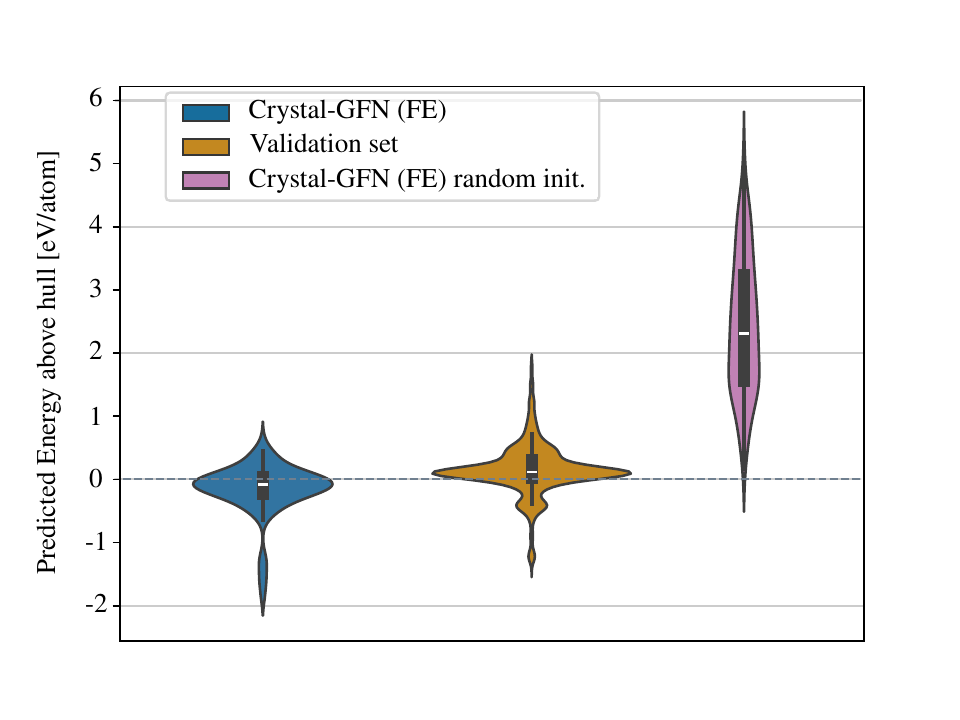}
    \caption{Distributions of the energy above hull, computed using the predicted formation energy w.r.t. to the Materials Project hull. The following samples are represented: in blue, samples from Crystal-GFN after training; in orange, the validation set, representative of the MatBench database; in pink, samples from an untrained Crystal-GFN. 65.6\% of the samples low below the convex hull, i.e. below zero, showing promising candidates.}
    \label{fig:ehull_distributions}
\end{figure}

\subsection{Diversity}

\begin{figure}
    \centering
    \includegraphics[width=0.85\textwidth]{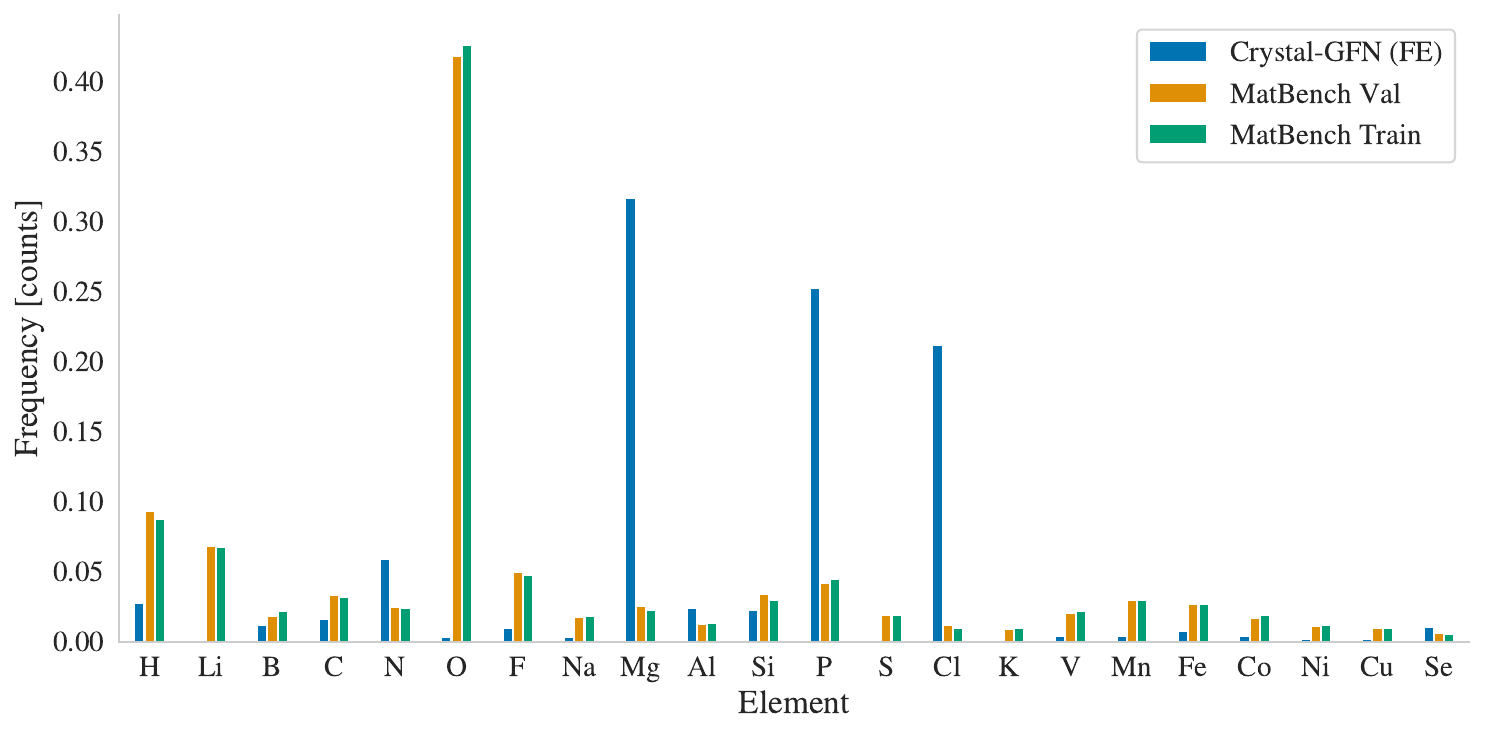}
    \caption{Distribution of total element prevalence per data set (10k FE Crystal-GFN samples, MatBench val, MatBench train). Each element is counted as per its stoichiometry in the crystal.}
    \label{fig:app:elements-counts}
\end{figure}

\begin{figure}
    \centering
    \includegraphics[width=0.85\textwidth]{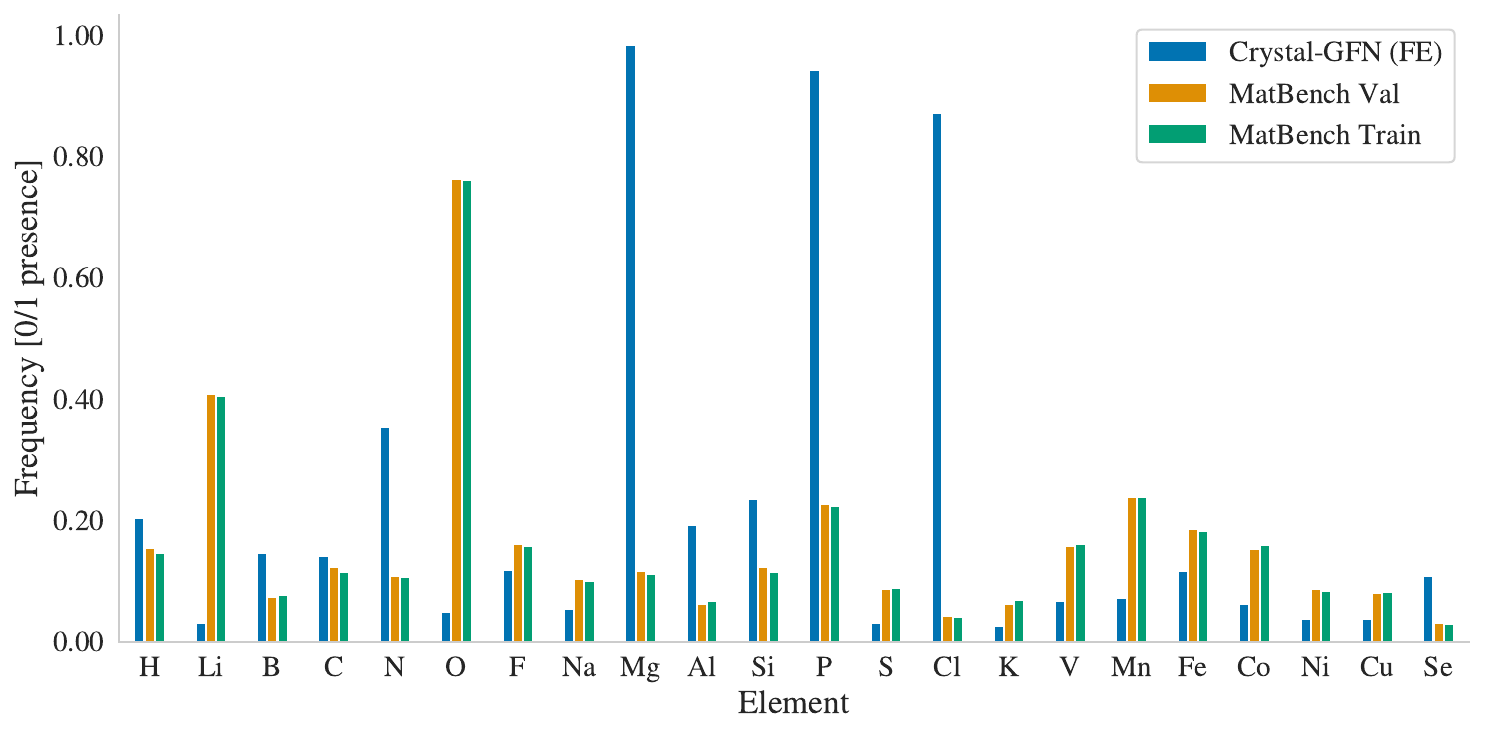}
    \caption{Distribution of binary element prevalence per data set (10k Crystal-GFlowNet samples, MatBench val, MatBench train). Each element is counted only once per crystal.}
    \label{fig:app:elements-counts-binary}
\end{figure}

\begin{figure}
    \centering
    \includegraphics[width=0.9\textwidth]{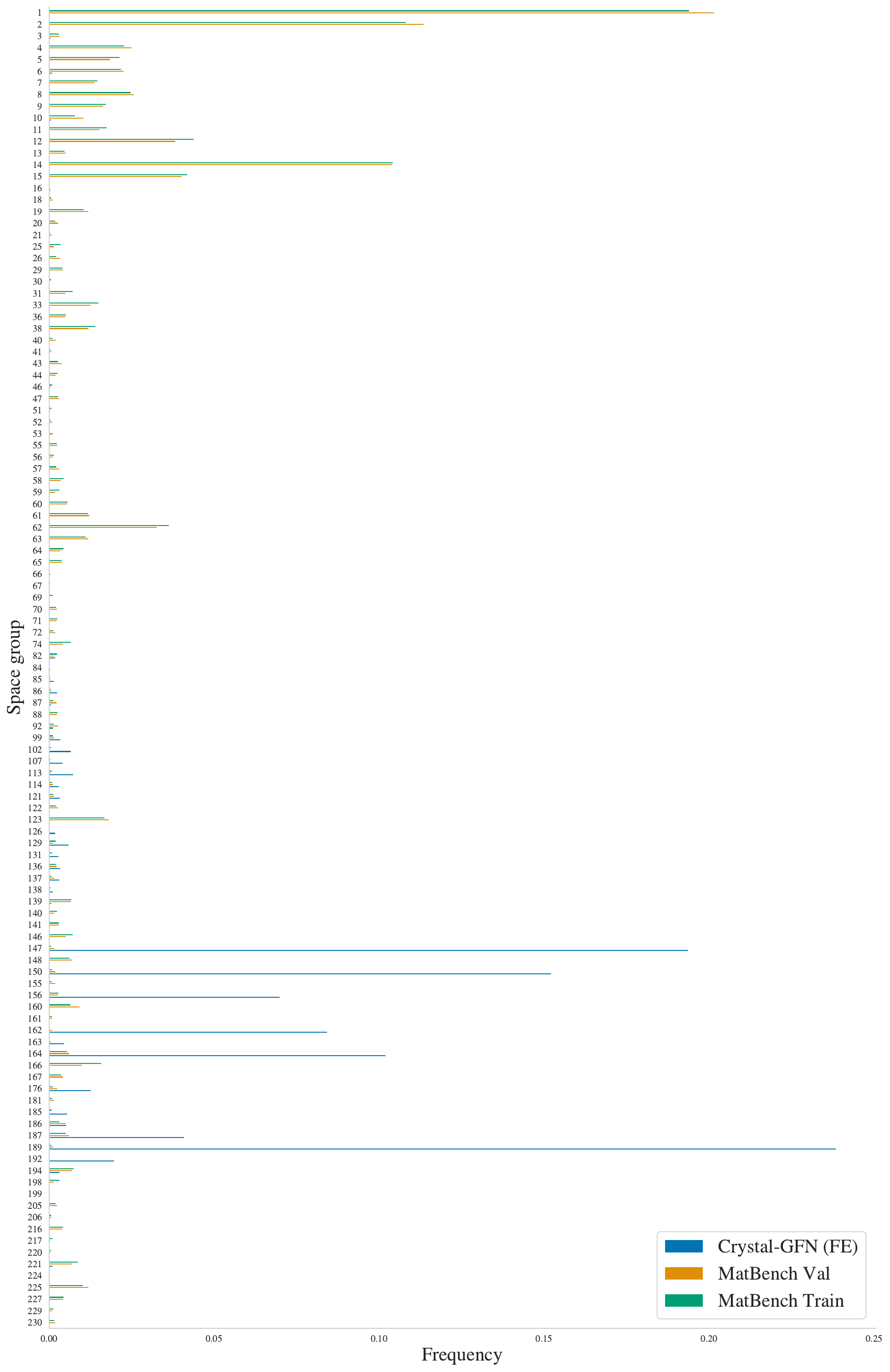}
    \caption{Distribution of space groups per data set (10k Crystal-GFN samples, MatBench val, MatBench train)}
    \label{fig:app:sg-counts}
\end{figure}

\begin{figure}
    \centering
    \includegraphics[width=0.85\textwidth]{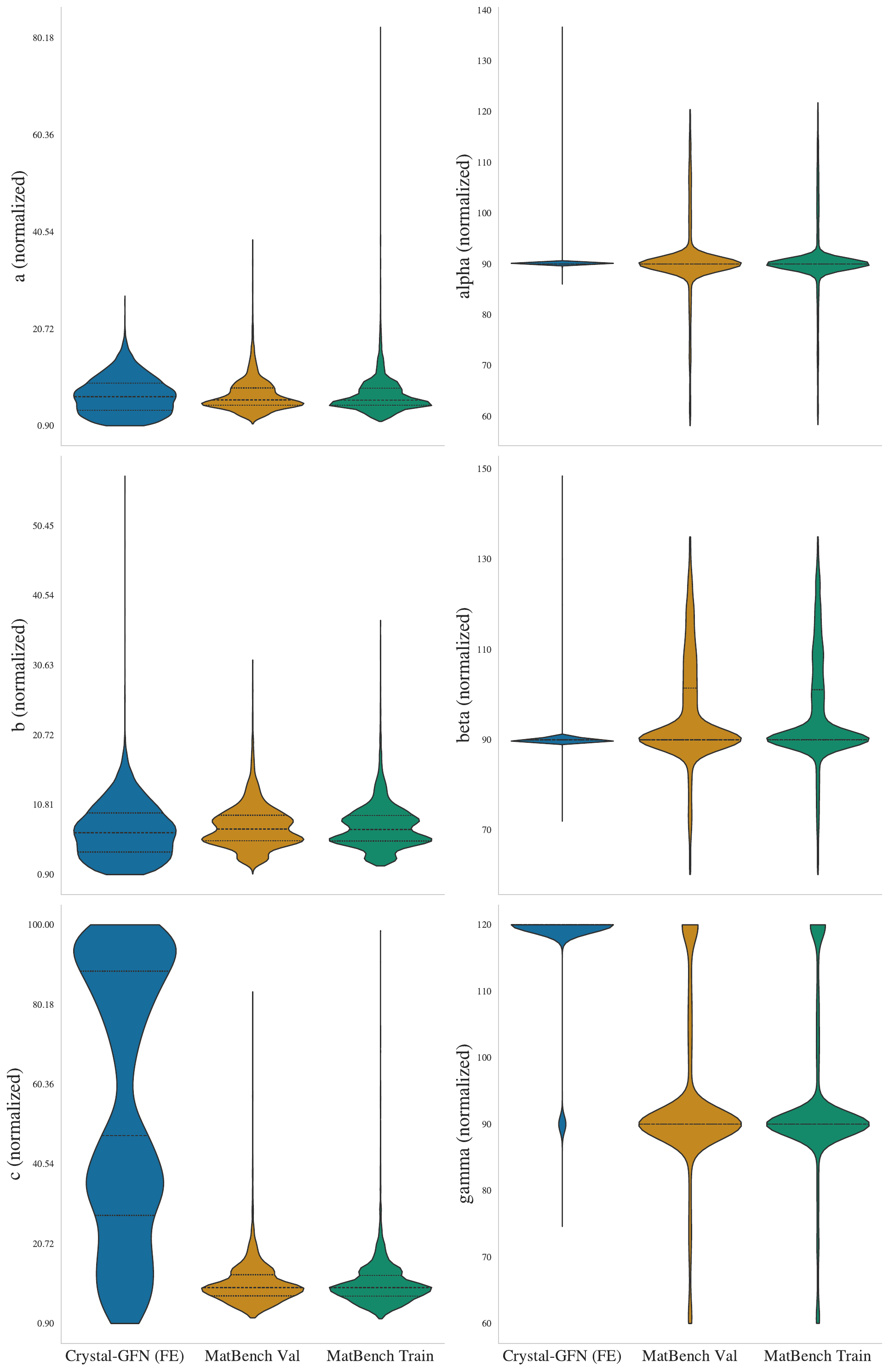}
    \caption{Distribution of lattice parameters sampled by Crystal-GFN (FE).}
    \label{fig:app:lattice-distrib}
\end{figure}

This section further demonstrates the diversity of the samples generated by the Crystal-GFN trained with an FE reward. They are displayed in \Cref{fig:app:elements-counts}, \Cref{fig:app:elements-counts-binary}, \Cref{fig:app:sg-counts} and \Cref{fig:app:lattice-distrib}. For simplicity we do not include their BG counterparts but observe similar, highly diverse results.

\begin{figure}
    \centering
    \includegraphics[width=0.85\textwidth]{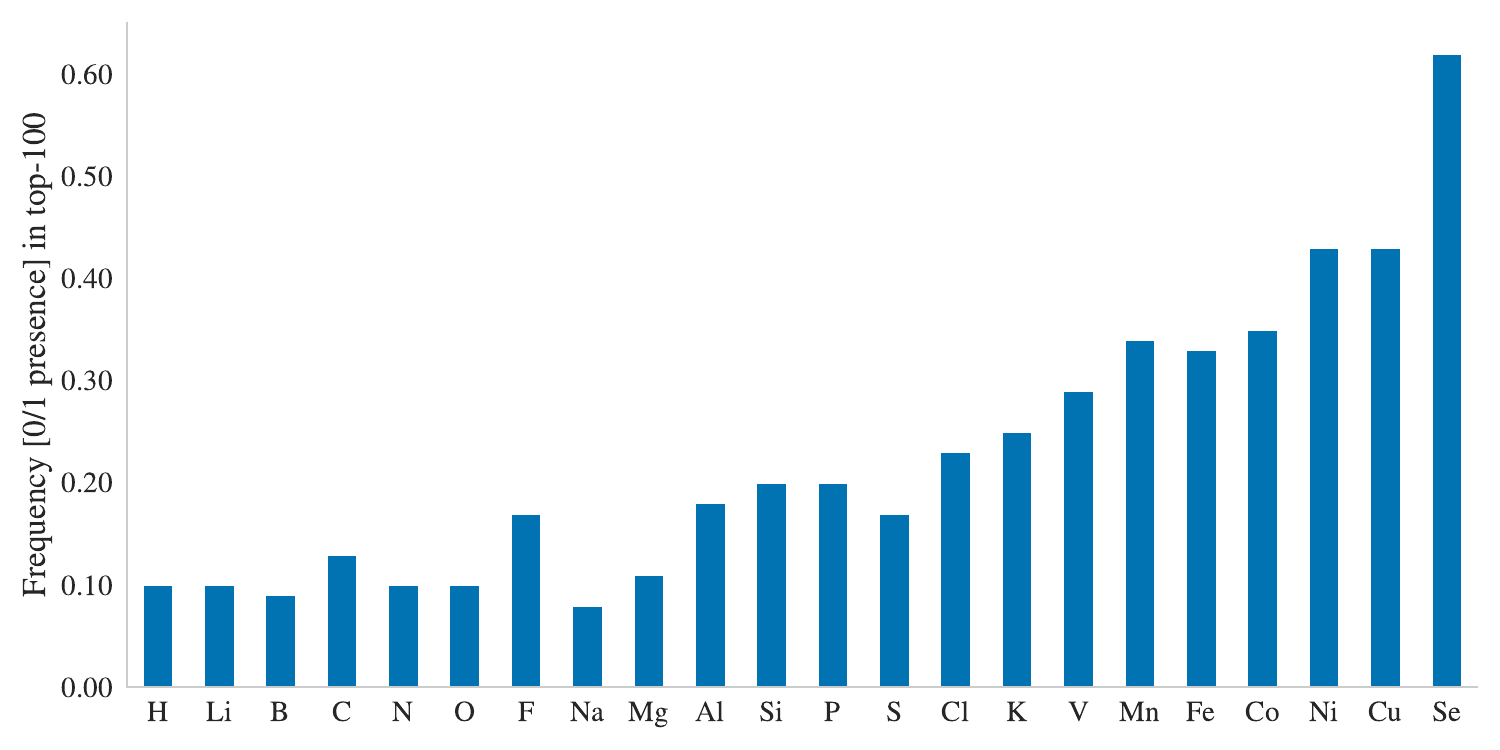}
    \caption{Distribution of element occurrences in the 100 most dense crystals sampled by Crystal-GFN (De).}
    \label{fig:app:top-dense-elems}
\end{figure}

\end{document}